\pdfoutput=1
\documentclass[11pt]{article}
\usepackage{acl}

\usepackage{times}
\usepackage{latexsym}

\usepackage[T1]{fontenc}
\usepackage[table]{xcolor}
\usepackage[utf8]{inputenc}
\usepackage{microtype}
\usepackage{stfloats}
\usepackage{todonotes}

\IfFileExists{inconsolata.sty}{\usepackage{inconsolata}}{}
\usepackage{booktabs} 
\usepackage{amsmath}
\usepackage{amssymb}
\usepackage{enumitem}
\usepackage{multirow}
\usepackage{graphicx}
\usepackage{array}
\usepackage{ragged2e}
\usepackage{tabularx}
\usepackage{placeins}
\graphicspath{{./}{../}{paper/}}
\usepackage[most]{tcolorbox}

\definecolor{typebg}{HTML}{E8F1FF}
\definecolor{substbg}{HTML}{EAF7EA}
\definecolor{scopebg}{HTML}{FFF3D6}
\definecolor{predbg}{HTML}{F3EAFE}
\definecolor{sevbg}{HTML}{FCE8E6}
\definecolor{intvbg}{HTML}{E6F4F1}
\newcommand{\hltype}[1]{\begingroup\setlength{\fboxsep}{1pt}\colorbox{typebg}{#1}\endgroup}
\newcommand{\hlsubst}[1]{\begingroup\setlength{\fboxsep}{1pt}\colorbox{substbg}{#1}\endgroup}
\newcommand{\hlscope}[1]{\begingroup\setlength{\fboxsep}{1pt}\colorbox{scopebg}{#1}\endgroup}

\newcommand{\hlintv}[1]{\begingroup\setlength{\fboxsep}{1pt}\colorbox{intvbg}{#1}\endgroup}

\newcolumntype{L}[1]{>{\RaggedRight\arraybackslash}p{#1}}
\newcolumntype{Y}{>{\RaggedRight\arraybackslash}X}

\title{LabGuard: Grounding Natural-Language Laboratory Rules\\ into Runtime Guards for Embodied Laboratory Agents}

\author{
  \textbf{Jingpu Yang\textsuperscript{1}\thanks{~~Equal contribution.}},
  \textbf{Fengxian Ji\textsuperscript{1,2}\footnotemark[1]},
  \textbf{Zhengzhao Lai\textsuperscript{3}\footnotemark[1]},
  \textbf{Zhexuan Cui\textsuperscript{3}}, \\
  \textbf{Guangxian Ouyang\textsuperscript{3}},
  \textbf{Qian Jiang\textsuperscript{3}},
  \textbf{Fan Zhang\textsuperscript{2}},
  \textbf{Min Peng\textsuperscript{1}}, \\
  \textbf{Qianqian Xie\textsuperscript{1}\thanks{Corresponding author.}},
  \textbf{Preslav Nakov\textsuperscript{2}},
  \textbf{Zhuohan Xie\textsuperscript{2}\footnotemark[2]} \\
  \textsuperscript{1}Wuhan University \quad
  \textsuperscript{2}MBZUAI \quad
  \textsuperscript{3}Northeastern University
  \\[0.35em]
    \texttt{\{fengxian.ji, fan.zhang, preslav.nakov, zhuohan.xie\}@mbzuai.ac.ae} \\
    \texttt{zhengzhaolai@cuhk.edu.cn, jingpuyang290@gmail.com, kasakura@outlook.com}\\
    \texttt{\{pengm, xieq\}@whu.edu.cn}, \texttt{\{202316187, 202219047\}@stu.neuq.edu.cn}
}

\begin{document}
\maketitle
\begin{abstract}
Scientific embodied agents are increasingly capable of carrying out laboratory procedures, but executing these procedures safely in dynamic laboratory environments remains challenging.
Current safety approaches often overlook the intermediate step of transforming laboratory natural language, including safety rules, manuals, protocols, and standard operating procedures, into machine-checkable runtime constraints.
We introduce \textbf{LabGuard}, Laboratory Guard, a language-to-execution safety suite that grounds natural-language laboratory rules into executable specifications and deploys them as runtime guards.
LabGuard includes three core components: \textbf{LabGuard-IR}, which defines a typed executable representation; 
\textbf{LabGuard-Bench}, which provides 812 supervised annotations expanded from 203 seed laboratory rules;
and \textbf{LabGuard-Grounder}, which maps natural-language laboratory rules into LabGuard-IR.
The resulting IR instances are handled by the \textbf{LabGuard Pipeline}, which compiles them into runtime monitors and applies them at the controller boundary.
Experiments show that LabGuard generalizes to unseen laboratory-rule sources, achieves 79.4 task-scope F1, and reduces unsafe events from 39.5\% to 23.8\% after monitor compilation.
In LabUtopia, its runtime monitors integrate with ACT, keeping interventions below 0.5\% while preserving task success.
\end{abstract}

\section{Introduction}
With the rapid development of foundation models, Vision-Language-Action (VLA) models, imitation-learning policies, and related techniques, embodied agents are moving beyond simple manipulation skills toward the execution of complex laboratory procedures.
Specifically, VLA models~\citep{rt2,openvla,pi0,meng} enhance robots' ability to generate actions from multimodal inputs; imitation-learning policies, such as ACT~\citep{act2023,Deng} and Diffusion Policy~\citep{diffusionpolicy2023,Jiang}, improve the learning and execution of laboratory manipulation skills; and laboratory simulation platforms and robotic chemistry systems, such as LabUtopia~\citep{labutopia2025}, RoboChemist~\citep{zhang2025robochemist}, and Organa~\citep{organa2024,ji2026finestate,lai2025can,luo2026natural,yang2024asynchronous,yang2025antijamming,yang2026frequency,cui2026textalign}, further establish scientific laboratories as an important evaluation setting for embodied agents.
However, in laboratory environments, task success does not necessarily imply safe execution, since agents must also follow safety knowledge expressed in natural-language rules, manuals, protocols, and standard operating procedures, covering chemical compatibility, equipment usage, procedural ordering, and risk intervention.

Existing safety research for laboratory embodied agents falls into three lines.
First, safety benchmarks test hazard recognition, risk assessment, unsafe-instruction refusal, and safety-aware planning, including LabSafety Bench, LABSHIELD, SafeAgentBench, VESTABENCH, and Safe-BeAl \citep{labsafetybench2024,sun2026labshield,safeagentbench2024,sadhu2025vestabench,huang2025framework,yang2024efficient}.
Second, decision-level methods improve planners or policies through safety-aware reasoning, alignment, and constrained learning, such as planning agents, risk-cognition modules, preference alignment, and constrained VLA optimization \citep{khan2025safety,yang2025antijamming}.
Third, runtime enforcement methods use guardrails, executable predicates, code-based monitors, or control constraints to block, replan, or restrict unsafe actions during execution \citep{wang2025robosafe,wang2025agentspec,ames2019control,ji2026servimage}.
However, these works often assume safety information is already formalized.
In laboratories, safety knowledge often appears as natural language, including SOPs, manuals, chemical-handling rules, and protocol instructions.
How to ground such text into executable monitor specifications for controller-side checking remains under-characterized.

The challenge is not simply to convert text into structured fields, but to make human-readable laboratory safety knowledge enter the robot control loop.
This requires a representable, learnable, and executable path from laboratory natural language to controller-side intervention, which raises three key questions.
First, how should laboratory safety language be represented so that rules can be activated and checked by a controller?
Second, how can models ground diverse sources, including SOPs, manuals, protocols, and chemical safety statements, into a typed executable representation?
Third, how can the grounded representation be compiled into runtime monitors that check live laboratory state before robot actions and pass, modify, or block actions when necessary?

\begin{figure*}[t]
    \centering
    \includegraphics[width=\textwidth]{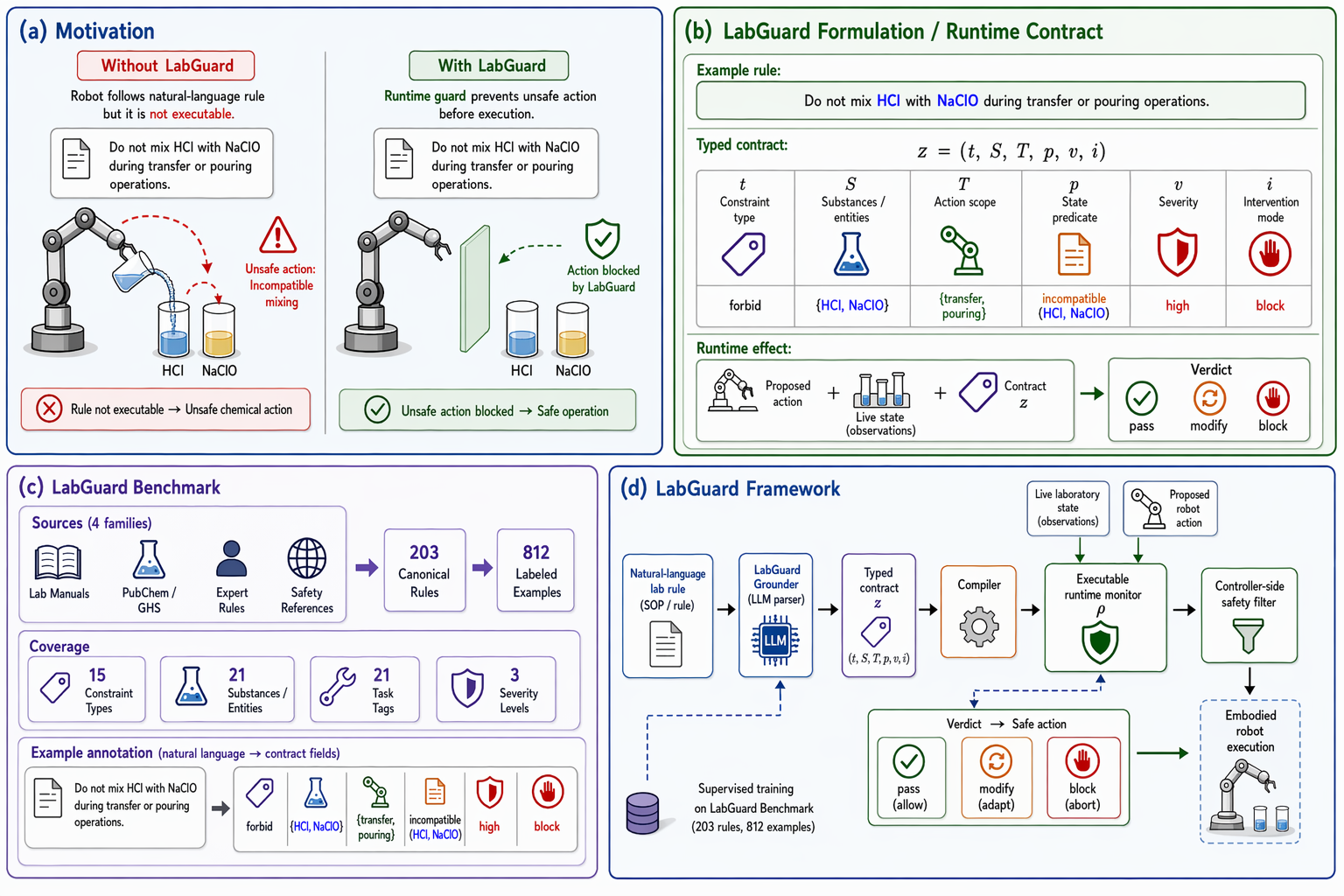}
    \caption{LabGuard Overview. Natural-language laboratory rules are grounded into LabGuard-IR, compiled into executable runtime monitors, and applied at the controller boundary during embodied laboratory execution.}    \label{fig:overview}
\end{figure*}

To address this problem, we propose \textbf{LabGuard}, a language-to-execution safety suite that transforms laboratory natural language into executable runtime guards for embodied laboratory agents.
LabGuard includes three core components for executable representation, supervised annotation, and laboratory-language grounding.
First, \textbf{LabGuard-IR} defines a typed executable representation, specifying the fields needed for runtime monitoring, including constraint type, entity binding, action scope, state predicates, severity, and intervention mode.
Second, \textbf{LabGuard-Bench} provides 812 supervised annotations for training and evaluation, enabling systematic assessment of how well models recover executable fields from laboratory natural language.
Third, \textbf{LabGuard-Grounder} maps natural-language laboratory rules into LabGuard-IR, turning human-readable safety knowledge into structured specifications that can be compiled.
The resulting IR instances are then handled by the \textbf{LabGuard Pipeline}, which compiles them into executable runtime monitors and executes these monitors at the controller boundary, where proposed actions can be passed, modified, or blocked according to the live laboratory state.
Experiments show that LabGuard generalizes to unseen laboratory-rule sources with 79.4 task-scope F1 and reduces unsafe events from 39.5\% to 23.8\% after monitor compilation.
In LabUtopia, its ACT-integrated monitors keep interventions below 0.5\% while preserving task success.

Our contributions follow the representable, learnable, and executable path introduced above.
First, we propose \textbf{LabGuard-IR}, a typed representation that makes laboratory safety language controller-usable by capturing constraint type, entity binding, action scope, state predicates, severity, and intervention mode.
Second, we construct \textbf{LabGuard-Bench}, which provides 812 supervised annotations expanded from 203 seed laboratory rules, and introduce \textbf{LabGuard-Grounder}, which maps natural-language laboratory rules into LabGuard-IR.
Third, we develop and evaluate the \textbf{LabGuard Pipeline}, which compiles grounded IR instances into runtime monitors and integrates them with LabUtopia + ACT control to test hazard reduction, task preservation, and intervention overhead.

\section{Related Work}
\textbf{LLM-based Embodied Agent Safety.}
Recent work on LLM-based embodied agent safety studies how agents recognize unsafe goals, refuse hazardous instructions, assess risk, or generate safety-aware plans~\citep{safeagentbench2024,agentsafe2025,safevla2025,Yang}. These studies typically evaluate safety at the instruction or planning level, where the model is asked to judge whether a goal, instruction, or proposed plan is safe before execution.
Related methods further improve decision-level safety through safety-aware reasoning, alignment, constrained learning, or risk-aware planning modules~\citep{safevla2025}.

\textbf{Runtime Assurance and Safety Filters for Robotics.}
Runtime assurance and safety-filtering methods aim to restrict unsafe behavior during execution by monitoring states, filtering actions, or enforcing formal constraints at the controller boundary.
Classical approaches include control barrier functions and related safety filters that constrain robot actions with respect to predefined safe sets~\citep{ames2019control,oscbf2025,zhang2026finreporting,zhang2026beyond,song2026maniplvm,yang2026geometric}.
Recent embodied-agent systems also use executable predicates, code-based monitors, visual monitors, or guardrail modules to inspect plans and actions before execution~\citep{codeasmonitor2025,wang2025robosafe,wang2025agentspec}.

\textbf{Natural Language Interfaces to Executable Systems.}
Semantic parsing maps natural language to executable formal representations such as SQL queries, robot commands, logical forms, and API calls~\citep{zelle1996learning,zettlemoyer2005learning,tellex2011understanding,matuszek2013learning}.
In scientific domains, prior work has extracted structured actions, materials, and experimental operations from synthesis procedures or laboratory protocols~\citep{mysore2019materials,vaucher2020automated}. Recent structured generation methods further improve output validity through lexical constraints, neural logic decoding, grammar-based parsing, and schema-constrained generation~\citep{hokamp2017lexically,lu2022neurologic,scholak2021picard,shin2021constrained,zhang2026alpha}.

\begin{table}[t]
\centering
\small
\setlength{\tabcolsep}{4pt}
\renewcommand{\arraystretch}{1.15}
\begin{tabular}{@{}p{0.30\columnwidth}p{0.60\columnwidth}@{}}
\toprule
\textbf{IR field} & \textbf{Grounded value} \\
\midrule
\multicolumn{2}{@{}p{0.94\columnwidth}@{}}{
\textbf{Natural-language laboratory rule:}
``\hlintv{Do not} \hltype{mix} \hlsubst{hydrochloric acid} with \hlsubst{sodium hypochlorite} during \hlscope{transfer} or \hlscope{pouring} operations.''
} \\
\midrule
\cellcolor{typebg}$\hat{t}$: constraint type
& \texttt{chemical\_incompatibility} \\
\cellcolor{substbg}$\hat{S}$: substances/entities
& \{\textit{HCl}, \textit{NaClO}\} \\
\cellcolor{scopebg}$\hat{T}$: action scope
& \{\textit{pour}, \textit{transfer}, \textit{mix}\} \\
\cellcolor{predbg}$\hat{p}$: state predicate
& Destination container must not already contain an incompatible substance. \\
\cellcolor{sevbg}$\hat{v}$: severity
& \texttt{high} \\
\cellcolor{intvbg}$\hat{i}$: intervention
& \texttt{block} \\
\midrule
\multicolumn{2}{@{}p{0.94\columnwidth}@{}}{
\textbf{Runtime effect:}
If the controller proposes a scoped transfer that would mix HCl and NaClO, the monitor blocks the action before execution.
} \\
\bottomrule
\end{tabular}
\caption{Example LabGuard-IR grounding. Colored spans in the laboratory rule correspond to the executable fields used to compile a runtime monitor.}
\label{tab:ir_example}
\end{table}

\section{LabGuard}
\label{sec:LabGuard}

\subsection{Problem Formulation}
\label{sec:problem_formulation}
Laboratory safety knowledge is often written in natural language for human practitioners, including SOPs, safety manuals, chemical handling rules, and protocol instructions, whereas runtime robot controllers require machine-checkable monitor specifications that can be activated, evaluated, and enforced during execution.
As illustrated in Figure~\ref{fig:overview}, this requires converting textual safety rules, such as chemical-incompatibility instructions, into typed executable specifications that can be checked against the current laboratory state and controller-proposed actions.
We formalize this as a laboratory-language-to-monitor transformation problem:
\begin{equation}
G_\theta(r)=z,\quad
E(\operatorname{compile}(z),s_t,a_t)=v_t.
\label{eq:sop_to_contract_problem}
\end{equation}
Here, $r$ denotes a natural-language laboratory rule, $G_\theta$ denotes the grounding model, $z$ denotes the predicted executable safety specification, $s_t$ denotes the live laboratory state, $a_t$ denotes the controller-proposed action, and $v_t \in \{\text{pass}, \text{modify}, \text{block}\}$ denotes the runtime safety verdict.
Thus, the goal is not to generate a textual explanation of a safety rule, but to recover the structured execution semantics needed by a runtime monitor.

\subsection{LabGuard-IR}
\label{sec:LabGuard_ir}

We define LabGuard-IR as the typed executable representation used as the target of laboratory-language grounding and the input to runtime monitor compilation.
It is designed to capture the execution semantics needed for controller-side safety checking, rather than only describing a rule in natural language.
Given a natural-language laboratory rule $r \in \mathcal{R}$, LabGuard-IR represents the grounded executable specification as
\begin{equation}
z = (\hat{t}, \hat{S}, \hat{p}, \hat{T}, \hat{v}, \hat{i}).
\label{eq:contract_ir}
\end{equation}
Here, $\hat{t}$ denotes the constraint type, $\hat{S}$ denotes the substance or entity set, $\hat{p}$ denotes the state or action predicate, $\hat{T}$ denotes the action scope, $\hat{v}$ denotes the severity level, and $\hat{i}$ denotes the intervention mode.
Table~\ref{tab:ir_example} provides a field-level example of this representation, corresponding to the typed executable view in Figure~\ref{fig:overview}(b).

These fields jointly make the grounded rule executable:
\begin{description}[leftmargin=2.0em, labelsep=0.5em, itemsep=0.1em, topsep=0.2em]
    \item[$\bullet$ $\hat{t}$] selects the safety logic to apply, such as incompatibility, temperature, equipment, or sequence checking.
    \item[$\bullet$ $\hat{S}$] links textual mentions to concrete laboratory entities that can be tracked in the environment.
    \item[$\bullet$ $\hat{p}$] defines the condition evaluated against the live state or proposed action.
    \item[$\bullet$ $\hat{T}$] activates the rule only for relevant controller actions.
    \item[$\bullet$ $\hat{v}, \hat{i}$] determine the response once a violation is detected, such as logging, modification, or blocking.
\end{description}

\subsection{LabGuard-Bench}
\label{sec:LabGuard-Bench}
LabGuard-Bench is a supervised benchmark for training and evaluating laboratory-language grounding, where each natural-language safety rule is aligned to an executable schema: constraint type, canonical substances, formal predicate, robot-task tags, severity, and intervention mode, so each annotation can be compiled into a runtime monitor.
The corpus contains \textbf{203 seed canonical rules} from four sources: laboratory safety manuals (71), PubChem GHS statements (62), expert-authored controller-facing rules (50), and chemical safety references (20), stratified by difficulty (easy 71, medium 105, hard 27). Seeds are expanded via paraphrase augmentation into \textbf{812 labeled examples}. The 15 constraint types group into four families: material compatibility, physical/quantitative handling, process/equipment checks, and environment/PPE/waste. Every seed rule and paraphrase was independently reviewed by all five annotators with simple majority adjudication ($\geq$3 of 5). Agreement distribution and review protocol are in Appendix~\ref{app:construction}.

% \todo{Please clarify the evaluation structure here. The paper says it has a three-layer evaluation, but then lists several protocols such as source-held-out, random 80/20, chemistry-held-out, stress replay, and LabUtopia runtime. Please explicitly separate the three layers: (1) language-to-contract grounding, (2) compiled-contract replay, and (3) live runtime integration.}
% \label{sec:eval}
%\todo[color=blue!25]{Incomplete?}

\subsection{LabGuard-Grounder}
\label{sec:LabGuard_grounder}

LabGuard-Grounder is the trainable component that maps natural-language laboratory rules into the LabGuard-IR defined in Eq.~\ref{eq:contract_ir}.
Given a supervised example $(r_i, z_i)$ from LabGuard-Bench, where $r_i$ is a natural-language laboratory rule and $z_i$ is the annotated executable specification, the grounder predicts $\hat{z}_i = G_\theta(r_i)$.
The predicted IR instance $\hat{z}_i$ follows the same schema as LabGuard-IR and is later passed to the compiler if it satisfies schema validation.

In our implementation, each IR instance $z_i$ is serialized as a structured text sequence, and the grounder is trained with supervised sequence generation.
The objective is the negative log-likelihood of the gold IR sequence:
\begin{equation}
\resizebox{0.89\linewidth}{!}{$
\mathcal{L}_{\mathrm{ground}}
=
-\sum_{i=1}^{N}
\sum_{m=1}^{|z_i|}
\log P_\theta(z_{i,m} \mid r_i, z_{i,<m}).
$}
\label{eq:grounder_loss}
\end{equation}
This objective trains the model to recover the executable fields required for downstream monitor compilation, including constraint type, substance or entity binding, state predicate, action scope, severity, and intervention mode.

At inference time, $\hat{z}_i$ is validated against the LabGuard-IR schema for required fields, categorical values, registry-resolved substances, and compiler-acceptable predicates.
Valid IR instances are passed to the compiler, while invalid outputs are rejected or flagged.
We implement the grounder with a LoRA fine-tuned language model and evaluate a hybrid variant with rule-based normalization for schema-sensitive fields such as substance mentions.

\begin{figure*}[!tbp]
    \centering
    \includegraphics[width=\textwidth]{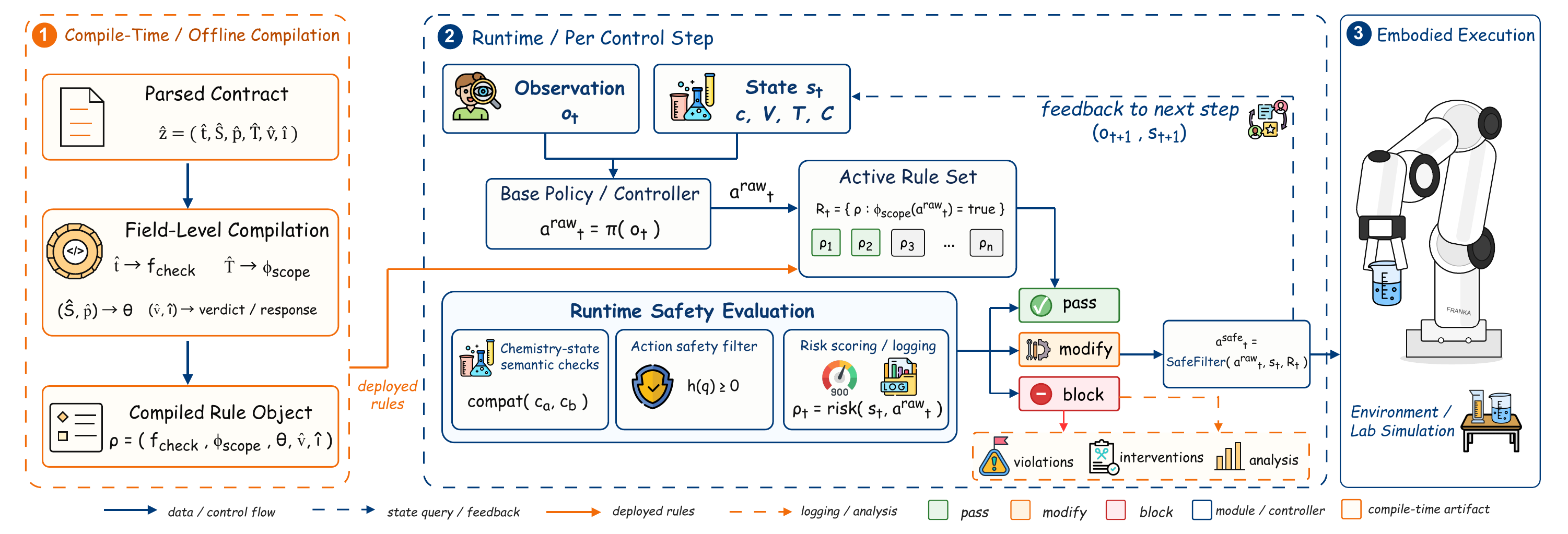}
    \caption{LabGuard Pipeline. Given a LabGuard-IR instance predicted by LabGuard-Grounder, the pipeline first compiles it into an executable runtime monitor and then executes the monitor at the controller boundary.}
    \label{fig:pipeline}
\end{figure*}

\section{LabGuard Pipeline}
\label{sec:LabGuard_pipeline}

Given a LabGuard-IR instance predicted by LabGuard-Grounder, the LabGuard Pipeline turns the structured representation into runtime safety behavior.
The pipeline consists of two stages: monitor compilation and runtime monitor execution.

\subsection{Monitor Compilation}
\label{sec:contract_compilation}

We instantiate the $\operatorname{compile}(\cdot)$ step introduced in Section~\ref{sec:problem_formulation}.
Given a grounded LabGuard-IR instance $\hat{z}$ predicted by LabGuard-Grounder, the compiler lowers it into an executable runtime monitor:
\begin{equation}
\rho = \operatorname{compile}(\hat{z})
= (f_{\mathrm{check}}, \phi_{\mathrm{scope}}, \theta, \hat{v}, \hat{i}).
\label{eq:compile_monitor}
\end{equation}
Here, $f_{\mathrm{check}}$ is a state-action check function, $\phi_{\mathrm{scope}}$ determines when the monitor is active, $\theta$ stores monitor parameters, $\hat{v}$ is the severity level, and $\hat{i}$ is the intervention mode.

The compiler constructs these monitor components from LabGuard-IR fields.
The constraint type selects $f_{\mathrm{check}}$ from a typed monitor library, e.g., mapping $\mathrm{chemical\_incompatibility}$ to $\operatorname{MaterialCompatibilityCheck}$.
The action scope instantiates $\phi_{\mathrm{scope}}$ so that the monitor is activated only for relevant controller actions.
The substance/entity set and predicate populate $\theta$ with registry-resolved substances, numeric thresholds, spatial bounds, or other check parameters.
The severity and intervention fields determine the logging priority and the runtime response once a violation is detected.

Before producing a monitor, the compiler validates schema conformance, resolves registry entries, and checks whether the IR instance can be lowered to an implemented monitor.
IR instances that fail these checks are rejected rather than executed.
All accepted monitors expose the same interface, allowing heterogeneous constraint types to be activated and evaluated by the same runtime execution loop.

\subsection{Runtime Contract Execution}
\label{sec:runtime_contract_execution}

LabGuard executes compiled monitors at the controller boundary and is agnostic to the underlying policy.
At each control step, a base controller $\pi$ proposes a raw action $a_t^{\mathrm{raw}} = \pi(o_t)$, and the runtime executor uses the live laboratory state $s_t$ to check whether this action satisfies the compiled monitors.
Here, $s_t$ denotes the live laboratory state, including per-container substance, volume, temperature, capacity, and robot kinematic state.

The executor first activates the monitors whose action scopes match the proposed action:
\begin{equation}
\mathcal{R}_t =
\{\rho \mid \phi_{\mathrm{scope}}^{\rho}(a_t^{\mathrm{raw}})=\mathrm{true}\}.
\end{equation}
Each active monitor then evaluates its check function against $(s_t, a_t^{\mathrm{raw}})$ and produces a verdict according to its intervention mode.
Given the raw action, live state, and active monitor set, the safety filter returns the executable action:
\begin{equation}
a_t^{\mathrm{safe}} =
\operatorname{SafeFilter}(a_t^{\mathrm{raw}}, s_t, \mathcal{R}_t).
\end{equation}
The safety filter may pass the raw action unchanged, modify it, or block it before execution; all interventions and violations are logged for analysis.

In our implementation, $\operatorname{SafeFilter}$ combines chemistry-state-aware semantic checks, a barrier-style action filter, and risk scoring.
The semantic checks evaluate IR predicates over live container states, including incompatibility, overflow, temperature, sequencing, PPE, and waste handling; the action filter enforces workspace, joint-margin, and velocity bounds; and the risk scorer flags high-risk steps from velocity, proximity, chemical-hazard, and uncertainty signals.
Full details are provided in Appendix~\ref{app:runtime_modules}.

\section{Experimental}
\label{sec:experiments}

\begin{table*}[t]
\centering
\small
\setlength{\tabcolsep}{3.5pt}
\begin{tabular}{@{}lcccccc@{}}
\toprule
\textbf{Method} & \textbf{Tag F1} & \textbf{Pred Soft} & \textbf{Pred EM} & \textbf{Unsafe Event $\downarrow$} & \textbf{False Interv. $\downarrow$} & \textbf{Task Success} \\
\midrule
\multicolumn{7}{c}{\emph{Source-held-out laboratory-language grounding (avg.\ 4 folds, mean$\pm$std):}} \\
\midrule
Regex BL & 58.7$\pm$9.2 & 28.5$\pm$12.3 & 15.8$\pm$9.7 & 34.2$\pm$8.5 & 8.3$\pm$3.2 & 72.4$\pm$7.8 \\
SciBERT & 54.9$\pm$10.8 & 14.2$\pm$7.1 & 7.6$\pm$4.8 & 41.7$\pm$9.3 & 12.5$\pm$4.7 & 68.3$\pm$8.9 \\
Qwen-7B 0-shot & 61.3$\pm$11.5 & 20.4$\pm$9.8 & 10.2$\pm$6.5 & 38.9$\pm$8.7 & 10.7$\pm$4.1 & 70.8$\pm$8.2 \\
Qwen3-32B 0-shot & 64.8$\pm$10.9 & 24.7$\pm$10.5 & 13.1$\pm$7.2 & 35.6$\pm$8.2 & 9.8$\pm$3.8 & 73.2$\pm$7.5 \\
DeepSeek-V3.1 0-shot & 67.2$\pm$10.3 & 27.3$\pm$11.1 & 14.9$\pm$7.8 & 33.8$\pm$7.9 & 9.2$\pm$3.5 & 74.6$\pm$7.1 \\
Qwen2.5-72B 0-shot & 68.9$\pm$9.8 & 29.1$\pm$11.6 & 16.3$\pm$8.3 & 32.4$\pm$7.6 & 8.7$\pm$3.3 & 75.8$\pm$6.8 \\
\midrule
\textit{Base Grounder (Qwen3-8B)} & 62.4$\pm$11.2 & 18.7$\pm$8.9 & 11.4$\pm$6.8 & 39.5$\pm$8.9 & 11.3$\pm$4.3 & 69.7$\pm$8.5 \\
\textit{+ LoRA fine-tuning} & 75.3$\pm$8.7 & 31.8$\pm$10.2 & 19.6$\pm$7.9 & 28.7$\pm$7.2 & 7.5$\pm$2.9 & 78.4$\pm$6.3 \\
\textit{+ Constraint decoding} & 77.8$\pm$8.2 & 34.2$\pm$9.8 & 21.3$\pm$7.5 & 26.3$\pm$6.8 & 6.9$\pm$2.6 & 79.8$\pm$5.9 \\
\textbf{LabGuard-Hybrid} & \textbf{79.4$\pm$7.9} & \textbf{36.7$\pm$9.5} & \textbf{23.1$\pm$7.2} & \textbf{23.8$\pm$6.4} & \textbf{6.2$\pm$2.4} & \textbf{81.2$\pm$5.6} \\
\midrule
\multicolumn{7}{c}{\emph{Chemistry-held-out laboratory-language grounding (hydrochloric acid, $N{=}9$, 5 seeds):}} \\
\midrule
Regex BL & 52.8$\pm$14.3 & 35.2$\pm$16.8 & 18.5$\pm$12.7 & 42.7$\pm$15.2 & 9.8$\pm$5.3 & 65.3$\pm$13.8 \\
Qwen2.5-72B 0-shot & 73.6$\pm$11.9 & 40.3$\pm$15.4 & 38.7$\pm$14.2 & 31.5$\pm$12.8 & 8.4$\pm$4.7 & 74.2$\pm$11.5 \\
\midrule
\textit{Base Grounder (Qwen3-8B)} & 68.4$\pm$13.5 & 32.8$\pm$16.2 & 28.3$\pm$14.8 & 36.9$\pm$13.7 & 10.5$\pm$5.1 & 70.8$\pm$12.9 \\
\textit{+ LoRA fine-tuning} & 88.7$\pm$9.2 & 68.4$\pm$12.8 & 52.3$\pm$15.1 & 18.6$\pm$9.4 & 5.7$\pm$3.8 & 86.4$\pm$8.7 \\
\textit{+ Constraint decoding} & 90.2$\pm$8.5 & 71.5$\pm$12.1 & 54.8$\pm$14.6 & 16.2$\pm$8.7 & 5.1$\pm$3.5 & 87.9$\pm$8.2 \\
\textbf{LabGuard-Hybrid} & \textbf{91.8$\pm$7.9} & \textbf{74.3$\pm$11.6} & \textbf{57.2$\pm$14.2} & \textbf{14.3$\pm$8.1} & \textbf{4.6$\pm$3.2} & \textbf{89.3$\pm$7.6} \\
\bottomrule
\end{tabular}
\caption{Laboratory-language grounding and compiled-monitor runtime outcomes under held-out settings. Grounding columns (Tag F1, Pred Soft, Pred EM) are evaluated on each held-out test set Appendix~\ref{app:eval_matrix}; runtime columns (Unsafe Event, False Interv., Task Success) aggregate LabUtopia L1--L4 episodes after monitor compilation Appendix~\ref{app:runtime_full}.}
\label{tab:nlp_results}
\end{table*}

\subsection{Experimental Setup}
\label{sec:exp_setup}
\paragraph{Benchmarks and evaluation settings.}
We evaluate LabGuard across three stages, using both our constructed benchmarks and existing runtime environments.
For laboratory-language grounding, we use LabGuard-Bench under source-held-out, random 80/20, and chemistry-held-out settings.
For compiled-monitor replay, we construct a 100-scenario stress replay set with 70 unsafe scenarios and 30 safe-control scenarios across 7 hazard categories.
For runtime monitor execution, we use LabUtopia Levels~1--4 with ACT as the base policy.
We additionally use a controlled pour scenario for module ablation.
Full benchmark and setting details are provided in Appendix~\ref{app:eval_matrix}, with runtime tasks and stress construction described in Appendix~\ref{app:runtime_tasks}.

\paragraph{Metrics.}
For laboratory-language grounding, we report Type Accuracy, Substance F1, Task Tag F1, Predicate Soft Match, Predicate Canonical EM, Schema Valid, and Compile Ready.
For compiled-monitor replay, we report check-function accuracy, task-scope F1, parameter recovery, catch rate, precision, and F1 on unsafe and safe-control scenarios.
For runtime monitor execution, we report Success Rate (SR), Violation Rate (VR), Intervention Rate (IR), Collision Rate (CR), mean Risk, Episode Length (EL), and Task Completion Time (TCT).
Detailed metric definitions are provided in Appendix~\ref{app:runtime_metrics}.

\paragraph{Baselines.}
For laboratory-language grounding, we compare against a Regex baseline, prompted LLMs, LoRA fine-tuned models, and our hybrid grounding variant.
For compiled-monitor replay, we compare monitors compiled from different grounding outputs, including Regex, LoRA, Hybrid, and gold labels / monitors compiled from gold labels.
For runtime monitor execution, we compare Base, which uses ACT~\citep{act2023} without the safety layer, against different runtime module configurations.

\subsection{Laboratory-Language Grounding}
\label{sec:grounding_results}

We first evaluate whether LabGuard can ground natural-language laboratory rules into executable IR instances under distribution shift.
Table~\ref{tab:nlp_results} reports source held-out and chemistry-held-out results.
LabGuard-Hybrid achieves the strongest performance, reaching 79.4 task-scope F1 under source-held-out evaluation and 91.8 under chemistry-held-out evaluation.
This indicates that LabGuard can still recover when safety rules should be activated even when source families or chemistry categories change.

LabGuard-Hybrid is effective because it combines learned task-scope prediction with rule-based substance normalization.
The learned component handles variation in laboratory-rule wording, while the rule-based component reduces substance-resolution errors that can prevent IR instances from compiling correctly.
Although exact predicate recovery remains challenging, the grounded IR instances reduce unsafe events from 39.5\% to 23.8\% after monitor compilation under source-held-out evaluation.
This suggests that accurate activation scope and reliable substance binding carry much of the downstream safety signal by routing proposed actions to the correct runtime checks.

\subsection{Compiled-Monitor Replay}
\label{sec:compilation_results}

\begin{figure*}[!t]
    \centering
\includegraphics[width=\textwidth]{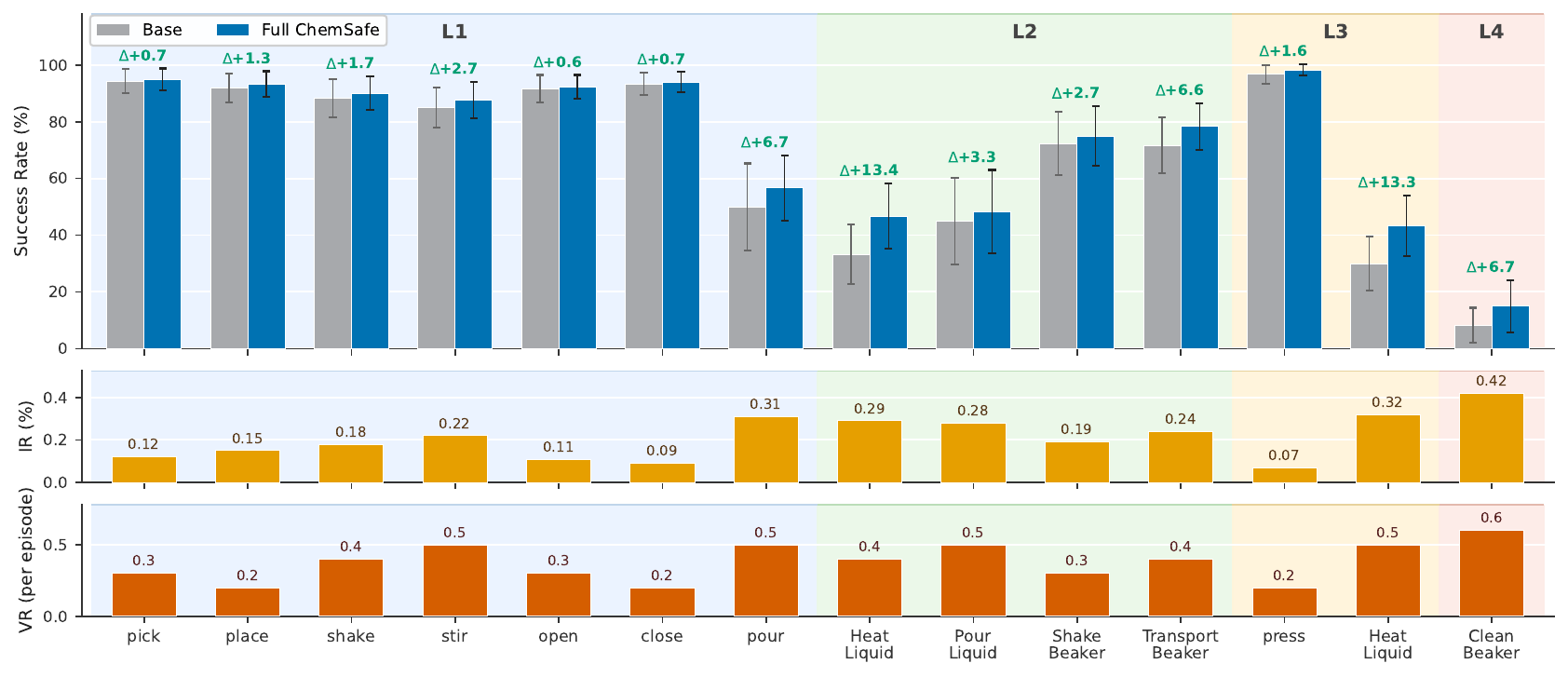}
    \caption{Per-task success rate (Base vs.\ Full LabGuard) across L1--L4. Mean SR over 3 seeds $\times$ 10 episodes (error bars: seed std), with IR and VR under Full and right-hand $\Delta$ values reporting Full $-$ Base.}
    \label{fig:per_task_runtime}
\end{figure*}

We next evaluate whether grounded IR instances can be compiled into deterministic monitors that prevent unsafe outcomes in offline replay.
The stress set contains 70 unsafe scenarios across 7 hazard categories and 30 safe-control scenarios; among the unsafe cases, 54 are in-schema and executable by the current monitor library, while 16 require additional monitor types such as exothermic-reaction tracking.
Gold IR labels catch all 54 in-schema unsafe scenarios, and LoRA and LabGuard-Hybrid approach this oracle bound.
The 16 out-of-schema cases are missed by all configurations, including gold, indicating a schema-coverage boundary rather than a translation failure.
Table~\ref{tab:source_held_out_propagation} further shows that source-held-out grounding preserves the relative ordering of methods after compilation, although cross-source degradation remains non-trivial.
\begin{table}[t]
\centering
\small
\setlength{\tabcolsep}{3.5pt}
\begin{tabular}{@{}lcccccc@{}}
\toprule
\textbf{Config.} & \textbf{TP} & \textbf{FN} & \textbf{FP} & \textbf{Prec} & \textbf{Catch} & \textbf{F1} \\
\midrule
No Safety & 7 & 63 & 1 & 87.5 & 10.0\% & 17.9 \\
Regex & 34 & 36 & 9 & 79.1 & 48.6\% & 60.2 \\
LoRA (Qwen3-8B) & 41 & 29 & 7 & 85.4 & 58.6\% & 69.5 \\
LabGuard-Hybrid & 41 & 29 & 7 & 85.4 & 58.6\% & 69.5 \\
Gold & 54 & 16 & 4 & 93.1 & 77.1\% & 84.4 \\
\bottomrule
\end{tabular}
\caption{Stress-benchmark results under compiled-monitor replay. Outcome-level catch, precision over the combined unsafe + safe-control pool, and their F1.
}
\label{tab:stress_benchmark}
\end{table}

\begin{table}[t]
\centering
\small
\setlength{\tabcolsep}{2pt}
\begin{tabular}{@{}lccccc@{}}
\toprule
\textbf{Grounding Module} & \textbf{Tag F1} & \textbf{Pred Soft} & \textbf{Catch\%} & \textbf{Prec} & \textbf{F1} \\
\midrule
\multicolumn{6}{c}{\emph{Source-held-out (avg.\ 4 folds):}} \\
\midrule
Regex & 45.1 & 33.3 & 47.1 & 77.8 & 58.7 \\
LoRA (Qwen3-8B) & 75.7 & 20.6 & 56.8 & 83.4 & 67.5 \\
LabGuard-Hybrid & 76.5 & 20.4 & 58.2 & 84.1 & 68.7 \\
\midrule
\multicolumn{6}{c}{\emph{Random 80/20 (upper bound):}} \\
\midrule
Regex & 68.3 & 54.2 & 48.6 & 79.1 & 60.2 \\
LoRA (Qwen3-8B) & 93.4 & 87.2 & 58.6 & 85.4 & 69.5 \\
LabGuard-Hybrid & 93.8 & 87.5 & 58.6 & 85.4 & 69.5 \\
\midrule
Gold & 100.0 & 100.0 & 77.1 & 93.1 & 84.4 \\
\bottomrule
\end{tabular}
\caption{Source-held-out propagation to stress benchmark. Same 70/30 pool as Table~\ref{tab:stress_benchmark}, re-grounded with source-held-out LoRA. 
}
\label{tab:source_held_out_propagation}
\end{table}

We analyze how grounding errors propagate through the LabGuard pipeline and affect downstream execution.
Grounding errors mainly concentrate in formal predicate composition, while task-scope grounding remains the most transferable signal under source shifts Table~\ref{tab:source_held_out_propagation}.
This pattern reflects three factors: schema complexity from 15 heterogeneous constraint types, sparse supervision with only 3--5 paraphrases per canonical rule, and metric limitations where token-level Soft Match may miss semantically equivalent predicates such as ``temp $>$ 80'' and ``80 $<$ temp''.

The compiled pipeline remains effective because runtime safety depends heavily on correct task-scope routing.
Once the system identifies when a rule should fire, deterministic check functions and registry-resolved substances recover much of the remaining execution semantics.
LabGuard-Hybrid further reduces substance-extraction failures by using regex fallback for the substance slot.
This explains why models with modest predicate scores can still yield measurable safety improvements after compilation.

\subsection{Runtime Monitor Execution}
\label{sec:runtime_monitor_execution}

The final stage closes the loop: compiled monitors must attach to learned controllers in a live simulator without degrading task success. We evaluate LabGuard on LabUtopia Levels~1--4 (per-task list in Appendix~\ref{app:runtime_tasks}; full numbers in Appendix~\ref{app:runtime_full}) 
to demonstrate runtime feasibility and bound the overhead of compiled-monitor execution.
The focus is on whether the safety layer can integrate with ACT policies without task-success loss; main-runtime VR/Unsafe Event are assurance-layer-logged (Appendix~\ref{app:runtime_metrics}), while an outcome-level oracle independent of the wrapper is applied only to the controlled pour ablation (Table~\ref{tab:ablation}).

Figure~\ref{fig:per_task_runtime} shows the safety layer is essentially a no-op on tasks already at ceiling and is associated with the largest task-success gains on safety-relevant operations. Intervention rates stay $\leq 0.5\%$ across tasks, and assurance-layer-logged violations remain in the $0.2$--$0.6$ range. This pattern confirms that compiled monitors do not introduce spurious interventions or degrade controller performance, completing the path from laboratory language to runtime monitoring without controller-side friction.

\subsection{Runtime Module Ablation}
\label{sec:module_ablation}

We isolate the contribution of each runtime module using a controlled \emph{pour} stress scenario that deliberately triggers unsafe behavior under a weak base policy.
This offline-rescored ablation uses an outcome-level oracle independent of the runtime assurance layer, and Table~\ref{tab:ablation} reports success rate (SR), intervention rate (IR), violation rate (VR), risk score, collision rate (CR), episode length (EL), and total cycle time (TCT).
Full LabGuard improves SR from 10\% to 40\% and reduces logged violations from 2.8 to 0.6 per episode.
Among individual modules, the Action Filter contributes the largest gain, raising SR to 30\% and reducing VR to 0.9, while Planning and Semantic modules provide complementary violation reductions.
Risk Scoring alone does not improve SR, but provides discriminative risk information for future adaptive interventions.
Overall, the ablation shows that runtime safety comes from the combination of task-scope routing, chemistry-state-aware checks, and kinematic filtering rather than any single module.

\begin{table}[t]
\centering
\small
\setlength{\tabcolsep}{2.2pt}
\begin{tabular}{@{}lccccccc@{}}
\toprule
\textbf{Configuration} & \textbf{SR$\uparrow$} & \textbf{IR} & \textbf{VR$\downarrow$} & \textbf{Risk} & \textbf{CR$\downarrow$} & \textbf{EL} & \textbf{TCT} \\
\midrule
Base (no safety) & 10\% & -- & 2.8 & 0.04 & 1.9 & 487 & 24.1s \\
Planning only & 20\% & 0.8\% & 2.1 & 0.05 & 1.6 & 512 & 25.3s \\
Semantic only & 10\% & 1.2\% & 1.5 & 0.05 & 1.3 & 495 & 24.8s \\
Action Filter only & 30\% & 2.1\% & 0.9 & 0.07 & 0.7 & 531 & 26.5s \\
Risk Scoring only & 10\% & 0.3\% & 2.5 & 0.12 & 2.1 & 476 & 23.7s \\
Full LabGuard & 40\% & 3.2\% & 0.6 & 0.08 & 0.5 & 548 & 27.4s \\
\bottomrule
\end{tabular}
\caption{Module-level ablation on the controlled \emph{pour} stress scenario. SR, IR, VR, Risk, CR, EL, TCT for each module in isolation and the full system.}
\label{tab:ablation}
\end{table}

\section{Conclusion}
We presented LabGuard, a language-to-execution safety suite that transforms laboratory natural language into typed executable specifications and runtime guards for controller-side intervention in simulation.
By separating laboratory-language grounding, monitor compilation, and runtime execution, LabGuard provides an auditable path from human-readable laboratory safety knowledge to executable monitors.
Through LabGuard-Bench and a three-stage evaluation protocol, we show that laboratory-language grounding and monitor compilation can link language-level safety rules to runtime robot behavior within a schema-scoped setting.
Future work should expand the monitor schema, improve cross-source predicate grounding, and evaluate stronger safety guarantees beyond simulation.

\section*{Limitations}
Our work is limited by its simulation-only evaluation. All runtime experiments and stress tests are conducted in the LabUtopia virtual environment, so the results show that LabGuard can connect laboratory natural language to runtime monitors in simulation, but they do not establish real-world laboratory safety. Physical deployment would require additional validation with real robot dynamics, sensing noise, calibration errors, and institutional safety procedures.
A second limitation is broader policy coverage. We use ACT as the main learned controller in LabUtopia, which is sufficient to test whether compiled monitors can be inserted at the controller boundary.
Future work can evaluate the same runtime-contract layer with more VLA models and manipulation policies to study how the safety layer behaves across different embodied agents.

\section*{Ethics Statement}

This work is conducted entirely in simulation and does not involve deployment on physical laboratory hardware or interaction with real hazardous chemicals. As a result, the experiments do not create direct real-world safety risks during evaluation. At the same time, the paper studies safeguards for scientific robotics, so we intentionally avoid overstating claims to reduce the risk of misleading conclusions about real-world laboratory safety.

\bibliography{custom}

\FloatBarrier
\appendix

\section{Evaluation Matrix}
\label{app:eval_matrix}

Table~\ref{tab:eval_matrix} summarizes the evaluation settings used in the main paper. We separate three kinds of evidence: laboratory-language grounding on held-out labels, offline replay of compiled monitors on scripted safety scenarios, and live LabUtopia execution with ACT. ``Offline replay'' means that compiled monitors are evaluated on scripted state traces; ``live runtime'' means that the safety layer runs inside the controller loop. This distinction is important because the stress benchmark uses outcome-level replay, whereas the L1--L4 runtime results report events logged by the active assurance layer.

\begin{table*}[t]
\centering
\footnotesize
\setlength{\tabcolsep}{3pt}
\resizebox{\textwidth}{!}{%
\begin{tabular}{@{}llllll@{}}
\toprule
\textbf{Protocol} & \textbf{Data \& Split} & \textbf{N} & \textbf{Primary Metric(s)} & \textbf{Grounding Ckpt} & \textbf{Mode} \\
\midrule
Grounding, random split & LabGuard-Bench, seed-42 80/20 split & 649 / 163 examples & Type Acc, Subst F1, Tag F1, Pred Soft/EM & per method & held-out labels \\
Grounding, source-held-out & LabGuard-Bench, leave-one-source-out & 4 folds & Tag F1, Pred Soft/EM & per method & held-out labels \\
Grounding, chemistry-held-out & HCl rules held out from training & 9 rules, 5 seeds & Tag F1, Pred Soft/EM & per method & held-out labels \\
Replay diagnostics & 41 held-out rules + scripted scenarios & 80 scenarios & Check Acc, Catch, Prec, Safety & per method & offline replay \\
Stress benchmark & 7 hazard categories + safe controls & 70 unsafe / 30 safe & TP, FN, FP, Catch, Prec, F1 & per method or gold & offline replay \\
\quad in-schema subset & stress cases covered by implemented monitors & 54 unsafe / 30 safe & Catch, Prec, F1 & per method or gold & offline replay \\
\quad out-of-schema subset & stress cases requiring new monitor types & 16 unsafe & missed by current schema & per method or gold & offline replay \\
Stress propagation & same stress pool re-grounded by held-out models & 70 unsafe / 30 safe & Tag F1, Pred Soft, Catch, Prec, F1 & source-held-out models & offline replay \\
Full LabGuard & L1--L4 tasks with ACT controller & 3 seeds $\times$ 10 episodes per task & SR, IR, VR, unsafe-event rate & full safety layer & live runtime \\
Module ablation & controlled adversarial \emph{pour} scenario & 3 seeds $\times$ 10 episodes & SR, IR, VR, Risk, CR, EL, TCT & module variants & live runtime + offline VR rescoring \\
\bottomrule
\end{tabular}%
}
\caption{Evaluation matrix. Summary of the grounding, offline replay, and live-runtime protocols used in the paper.}
\label{tab:eval_matrix}
\end{table*}

\section{Runtime Safety Module Details}
\label{app:runtime_modules}

This appendix gives additional implementation details for the runtime monitor execution modules summarized in Section~\ref{sec:runtime_monitor_execution}. The modules operate after laboratory-language grounding and monitor compilation. No language model is invoked inside the controller loop.

\subsection{Compiled Monitor Target}

Given a grounded LabGuard-IR instance,
\[
\hat{z}=(\hat{t},\hat{S},\hat{p},\hat{T},\hat{v},\hat{i}),
\]
the compiler lowers it into a runtime monitor,
\[
\rho=\mathrm{compile}(\hat{z})=(f_{\mathrm{check}},\phi_{\mathrm{scope}},\theta,\hat{v},\hat{i}).
\]
Here, $f_{\mathrm{check}}$ is the state-action check function, $\phi_{\mathrm{scope}}$ determines whether the monitor is active for a proposed action, $\theta$ stores monitor parameters, $\hat{v}$ is the severity level, and $\hat{i}$ is the intervention mode. The scope predicate is instantiated from the action-scope field:
\[
\phi_{\mathrm{scope}}(a)=\mathbf{1}[\mathrm{tag}(a)\in \hat{T}].
\]
The constraint type $\hat{t}$ selects a monitor class from the typed monitor library, while $\hat{S}$ and $\hat{p}$ populate $\theta$ with registry-resolved substances, thresholds, spatial bounds, or other check parameters.

For material-compatibility specifications, the monitor uses an incompatibility table over registered substances. For two substances $c_a$ and $c_b$, the lookup is
\begin{equation}
\mathrm{compat}(c_a,c_b)=
\begin{cases}
1, & c_b\notin\mathrm{inc}(c_a)\wedge c_a\notin\mathrm{inc}(c_b),\\
0, & \mathrm{otherwise},
\end{cases}
\end{equation}
where $\mathrm{inc}(\cdot)$ denotes the incompatible-substance list in the runtime registry. This lookup is intentionally simple. Violations are triggered by explicit registry entries and task-conditioned monitor templates, not by a symbolic planner or reaction simulator.

\subsection{Chemistry-State-Aware Semantic Checks}

The runtime assurance layer maintains a live laboratory state containing registered containers, their substance identities, volumes, temperatures, capacities, and the robot kinematic state. Active monitors evaluate their predicates against this state and the controller-proposed action.

For transfer-like actions, the monitor inspects the source container, destination container, transferred substance, destination contents, and planned transfer volume. The supported checks include incompatible mixing, destination overflow, temperature-threshold violations, sequencing preconditions, PPE-related preconditions, and waste-handling constraints. If a violation is detected, the monitor returns a verdict according to its intervention mode. The action may be blocked or modified before execution, and the intervention is logged for analysis. If no violation is detected, the state update is committed so that later monitors operate on the updated container state.

\subsection{Geometric and Kinematic Bounds}

The runtime layer also applies lightweight geometric and kinematic validation. These checks enforce fixed workspace bounds, conservative joint margins, and per-joint velocity limits. The workspace constraint is applied componentwise:
\begin{equation}
p_{\min,k}\leq p_{ee,k}\leq p_{\max,k}, \qquad k\in\{x,y,z\},
\end{equation}
where $p_{ee}$ is the end-effector position. The action-side velocity constraint is
\begin{equation}
|\dot{q}_j|\leq \alpha \dot{q}_{j,\max}, \qquad \alpha=0.8.
\end{equation}
These bounds are fixed in the current implementation and are used as conservative runtime safeguards rather than learned safety constraints.

\subsection{Barrier-Style Action Filter}

The action filter uses barrier-style margins around joint limits together with velocity clamping. For joint $j$, the margins are
\begin{align}
h_j^{+}(q) &= q_{j,\max}-q_j-\delta,\\
h_j^{-}(q) &= q_j-q_{j,\min}-\delta,
\end{align}
where $\delta$ is a fixed safety margin. These quantities are used to steer target positions away from joint boundaries and to clamp velocities. Although the design is inspired by control-barrier constraints, the implementation does not solve a quadratic program and does not claim the guarantees of a full optimization-based CBF controller.

\subsection{Risk Scoring and Logging}

The risk scorer aggregates velocity, proximity, chemical hazard, and uncertainty with fixed hand-designed weights:
\begin{equation}
\rho_t = 0.25\rho_v + 0.25\rho_p + 0.30\rho_c + 0.20\rho_u.
\end{equation}
The coefficients are fixed heuristics rather than learned parameters. When chemistry-state-aware checks are enabled, the chemical term $\rho_c$ is computed from the live container state, including substance identities, volumes, fill fractions, and temperature-related signals. High-risk steps are logged for analysis, and threshold exceedances can produce safety events according to the active monitor configuration. In the current implementation, risk scoring is primarily diagnostic: it flags high-risk states and supports intervention analysis, but it does not by itself provide a formal safety guarantee or restore simulator state.

\section{Runtime Metric Definitions}
\label{app:runtime_metrics}

We report runtime metrics under two scoring modes. For L1--L4 LabUtopia runs, safety events are logged by the runtime assurance layer when that layer is active. For the controlled \emph{pour} ablation, we additionally rescore each trajectory with an offline compiled-monitor checker, so that the Base policy, which has no active safety layer, can be compared with the safety-module variants.

\paragraph{Success Rate.}
SR is the fraction of episodes that complete the task. Higher SR indicates better task performance.

\paragraph{Violation Rate.}
VR is the average number of safety violations per episode. In the L1--L4 runtime table, VR counts violations logged by the active assurance layer. In the controlled \emph{pour} ablation, VR is computed by replaying each trajectory through the same compiled-monitor checker. This offline scoring makes the Base row comparable to rows with an active safety layer.

\paragraph{Intervention Rate.}
IR is the fraction of controller steps in which the safety layer blocks, modifies, or replaces the proposed action. It measures how often the runtime layer changes controller behavior. Lower IR is preferable when safety outcomes are comparable.

\paragraph{Unsafe Events and False Interventions.}
Unsafe Event is reported for runtime-style summaries as the percentage of episodes in which the scoring protocol observes an unsafe outcome. In offline replay settings, the analogous outcome-level quantities are reported as catch, precision, and F1. False Intervention measures interventions on safe-control cases or safe execution segments, and is reported only when such safe-control labels are available.

\paragraph{Risk, Collision, and Runtime Cost.}
Risk is the mean heuristic risk score from the runtime scorer. CR is the average number of collision events per episode in the controlled ablation. EL is the episode length in control steps, and TCT is the total cycle time. These quantities are reported in the controlled ablation to describe side effects of each runtime module.

\section{LabGuard-Bench Construction Protocol}
\label{app:construction}

\subsection{Annotation Protocol}

The corpus starts from 203 seed rules from four source families described in Section~\ref{sec:LabGuard-Bench}. Each seed is labeled with constraint type, canonical substances, predicate, task tags, severity, and intervention mode. Paraphrases inherit the seed label only after review for semantic equivalence.

Five annotators with chemistry or robotics background reviewed the seed labels and paraphrases. Final labels use simple majority agreement. Items without majority agreement are revised before inclusion. This protocol is intended to keep paraphrase augmentation from adding or removing safety constraints.

We keep source-family metadata for each item and do not rely on verbatim source text in the released labels. Table~\ref{tab:per_type_counts} reports the final distribution over the 15 constraint types. The counts sum to 812 examples. Per-source seed counts are 71 lab-manual rules, 62 PubChem GHS statements, 50 expert-authored controller-facing rules, and 20 chemical-safety reference rules.

\begin{table}[t]
\centering
\small
\setlength{\tabcolsep}{4pt}
\begin{tabular}{@{}lr@{\hspace{1.5em}}lr@{}}
\toprule
\textbf{Constraint type} & \textbf{N} & \textbf{Constraint type} & \textbf{N} \\
\midrule
chemical\_incompatibility & 132 & velocity\_limit & 40 \\
equipment\_check & 100 & waste\_handling & 36 \\
sequence\_constraint & 80 & spill\_prevention & 36 \\
exposure\_limit & 68 & container\_capacity & 24 \\
temperature\_constraint & 64 & storage\_constraint & 24 \\
ventilation\_requirement & 60 & workspace\_zone & 20 \\
fire\_safety & 56 & force\_limit & 20 \\
ppe\_requirement & 52 & \textbf{Total} & \textbf{812} \\
\bottomrule
\end{tabular}
\caption{\textbf{Per-constraint-type distribution} across the 812-example LabGuard-Bench corpus.}
\label{tab:per_type_counts}
\end{table}

For source-held-out evaluation, all seed rules and paraphrases from the held-out source family are removed from training. Chemical names and constraint types can still appear across folds because they belong to the shared schema and registry. The split is therefore designed to test source phrasing and rule-structure shift, not unseen chemistry vocabulary alone.

\begin{table*}[t]
\centering
\small
\setlength{\tabcolsep}{4pt}
\begin{tabularx}{\textwidth}{@{}L{0.17\textwidth}L{0.24\textwidth}L{0.27\textwidth}Y@{}}
\toprule
\textbf{Source} & \textbf{Language Style / Safety Semantics} & \textbf{Dominant Constraint Families} & \textbf{Main Compiled Checks} \\
\midrule
Expert-authored & Controller-facing laboratory instructions for robot manipulation, transfer, and device use & Equipment checks, sequencing, chemical compatibility, temperature limits & Sequence enforcement, material compatibility, heat-safety checks \\
PubChem GHS & Short hazard / precaution statements about exposure, PPE, storage, and fire risk & Exposure, ventilation, PPE, fire safety, waste handling & Motion-limit, heat-safety, and procedure-order checks \\
Chemical safety reference & Curated pairwise incompatibility knowledge and reagent interaction rules & Material compatibility & Material-compatibility checks \\
Lab safety manual & Procedure-focused laboratory rules for transfer, cleanup, workspace, and device interaction & Sequencing, temperature, spill prevention, equipment checks, velocity limits & Sequence enforcement, heat-safety, pour-speed, and motion-limit checks \\
\bottomrule
\end{tabularx}
\caption{Source families in LabGuard-Bench. Each family contributes rules that map to one or more implemented monitor types.}
\label{tab:construction}
\end{table*}

\section{Runtime Evaluation Details}
\label{app:runtime_tasks}
\label{app:runtime_full}
\label{app:act_training}
\label{app:act_details}

The ACT base policy follows the standard ACT setup used in the main runtime experiments. It uses two RGB camera views at $256{\times}256$ resolution and robot state, and each task model is trained from roughly 50 scripted demonstrations. This small demonstration budget is one reason the runtime results have high variance on long-horizon or safety-sensitive tasks.

We evaluate on LabUtopia Levels 1--4. Level 1 includes pick, place, press, shake, stir, open, close, and pour. Level 2 includes HeatLiquid, PourLiquid, ShakeBeaker, and TransportBeaker. Level 3 uses OOD versions of press and HeatLiquid. Level 4 uses CleanBeaker. Each task is evaluated with 3 random seeds and 30 episodes per seed. Table~\ref{tab:runtime_main_full} gives the per-task results.

\begin{table*}[t]
\centering
\small
\setlength{\tabcolsep}{8pt}
\begin{tabular}{@{}lcccccc@{}}
\toprule
\textbf{Task} & \textbf{Base SR} & \textbf{Full SR} & \textbf{$\Delta$SR} & \textbf{Full IR (\%)} & \textbf{Full VR} & \textbf{Notes} \\
\midrule
L1 pick & 94.3$\pm$4.2 & 95.0$\pm$3.8 & +0.7 & 0.12\% & 0.3 & ceiling \\
L1 place & 92.0$\pm$5.1 & 93.3$\pm$4.5 & +1.3 & 0.15\% & 0.2 & ceiling \\
L1 shake & 88.3$\pm$6.7 & 90.0$\pm$5.9 & +1.7 & 0.18\% & 0.4 & modest \\
L1 stir & 85.0$\pm$7.2 & 87.7$\pm$6.4 & +2.7 & 0.22\% & 0.5 & modest \\
L1 open & 91.7$\pm$4.8 & 92.3$\pm$4.2 & +0.6 & 0.11\% & 0.3 & ceiling \\
L1 close & 93.3$\pm$3.9 & 94.0$\pm$3.5 & +0.7 & 0.09\% & 0.2 & ceiling \\
L1 pour & 50.0$\pm$15.3 & 56.7$\pm$11.5 & +6.7 & 0.31\% & 0.5 & improvement \\
\midrule
L2 HeatLiquid & 33.3$\pm$10.4 & 46.7$\pm$11.5 & +13.4 & 0.29\% & 0.4 & improvement \\
L2 PourLiquid & 45.0$\pm$15.3 & 48.3$\pm$14.7 & +3.3 & 0.28\% & 0.5 & modest \\
L2 ShakeBeaker & 72.3$\pm$11.2 & 75.0$\pm$10.5 & +2.7 & 0.19\% & 0.3 & modest \\
L2 TransportBeaker & 71.7$\pm$9.8 & 78.3$\pm$8.2 & +6.6 & 0.24\% & 0.4 & improvement \\
\midrule
L3 press & 96.7$\pm$3.2 & 98.3$\pm$2.1 & +1.6 & 0.07\% & 0.2 & ceiling \\
L3 HeatLiquid & 30.0$\pm$9.5 & 43.3$\pm$10.8 & +13.3 & 0.32\% & 0.5 & improvement \\
\midrule
L4 CleanBeaker & 8.3$\pm$6.1 & 15.0$\pm$9.2 & +6.7 & 0.42\% & 0.6 & long-horizon \\
\bottomrule
\end{tabular}
\caption{Full per-task runtime results across L1--L4 tasks. We report success rate (SR, mean$\pm$std over 3 seeds), $\Delta$SR (Full LabGuard minus Base), intervention rate (IR), and \emph{assurance-layer-logged} violations per episode (VR; see Appendix~\ref{app:runtime_metrics} for VR semantics) under Full LabGuard, with a qualitative note column summarizing per-task behavior.}
\label{tab:runtime_main_full}
\end{table*}

\end{document}